%% file: main.tex
\documentclass[aic,crcready]{iosart2x}
\usepackage{bm}
\usepackage{graphicx}

\DeclareRobustCommand{\[}{\begin{equation}}
\DeclareRobustCommand{\]}{\end{equation}}
\newcommand{\newterm}[1]{{\it #1}}

\let\citep\cite
\let\citet\cite

\def\ve{{\bm{e}}}

\def\vx{{\bm{x}}}

\def\sS{{\mathbb{S}}}

\newcommand{\heart}{\ensuremath\heartsuit}

\pubyear{0000}
\volume{0}
\firstpage{1}
\lastpage{1}

\begin{document}

\begin{frontmatter}


\title{Token-Modification Adversarial Attacks for Natural Language Processing: A Survey}
\runtitle{Token-Modification Adversarial Attacks}











\begin{aug}
\author[A,B]{\inits{T.}\fnms{Tom} \snm{Roth}\ead[label=e1]{thomas.p.roth@student.uts.edu.au}}
\author[B]{\inits{Y.}\fnms{Yansong} \snm{Gao}\ead[label=e2]{garrison.gao@data61.csiro.au}}
\author[B]{\inits{A.}\fnms{Alsharif} \snm{Abuadbba}\ead[label=e3]{sharif.abuadbba@data61.csiro.au}}
\author[B]{\inits{S.}\fnms{Surya} \snm{Nepal}\ead[label=e4]{surya.nepal@data61.csiro.au}}
\author[A]{\inits{W.}\fnms{Wei} \snm{Liu}\ead[label=e5]{wei.liu@uts.edu.au}}
\address[A]{\orgname{University of Technology Sydney}, NSW, \cny{Australia}\printead[presep={\\}]{e1,e5}}
\address[B]{\orgname{CSIRO's Data61}, Sydney, NSW, \cny{Australia}\printead[presep={\\}]{e2,e3,e4}}
\end{aug}

\begin{abstract}
Many adversarial attacks target natural language processing systems, most of which succeed through modifying the individual tokens of a document. Despite the apparent uniqueness of each of these attacks, fundamentally they are simply a distinct configuration of four components: a goal function, allowable transformations, a search method, and constraints. In this survey, we systematically present the different components used throughout the literature, using  an attack-independent framework which allows for easy comparison and categorisation of components. Our work aims to serve as a comprehensive guide for newcomers to the field and to spark targeted research into refining the individual attack components.

\end{abstract}


\end{frontmatter}

\section{Introduction}
Adversarial machine learning investigates the susceptibility of machine learning models to manipulations by an adversary through an \textit{adversarial attack}~\citep{Kurakin2017Adversarial}. While the majority of adversarial machine learning research has focused on image processing, there has been rapidly increasing interest in the natural language processing (NLP) domain, motivated by the unique challenges and opportunities presented by textual data. In this work, we survey the recent advancements in adversarial attacks within the NLP domain. 

\begin{figure}[!h]
\centering
\includegraphics[width=0.75\linewidth]{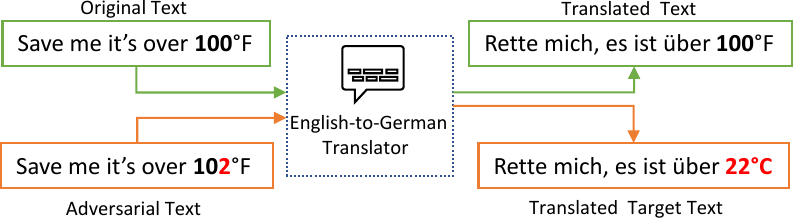}
\caption{Example of a token-modification adversarial attack on a machine translation model. 
}
\label{fig:attack_demo}
\end{figure}

An \textit{adversarial example} is a carefully crafted instance created by introducing a perturbation to an \textit{original example}. This perturbation is designed in such a way that the attacked \textit{victim model} processes the original example as expected but fails on the adversarial example in a manner predetermined by the attacker. To illustrate this concept, \citet{Wallace2020Imitation} have aimed to perturb a document reading \textit{``Save me it's over 100F''} to introduce a specific token \textcolor{blue}{``\textit{22C}''} into the output of an English-to-German translation model (see Figure \ref{fig:attack_demo}). Their attack algorithm achieved this by replacing \textit{``100F''} with \textit{``10\textcolor{blue}{2}F''}. While the model translates the original example correctly as \textit{``Rette mich, es ist über 100F''}, it now translates the adversarial example as \textit{``Rette mich, es ist über \textcolor{blue}{22C}''}, thereby introducing the specific token, and in turn, effectively demonstrating the potential potency of adversarial attacks. 

Usually text adversarial examples must also meet a set of constraints. For example, one constraint might specify that the semantic meaning of the original example is maintained. In this case,  \textit{``Save me it's \textcolor{blue}{above} 100F''} would be valid, but not the proposed \textit{``Save me it's over 10\textcolor{blue}{2}F''}. A second constraint might require that the adversarial example appear non-suspicious to a human. This rules out \textit{``Save \textcolor{blue}{MySeLFFFFF} it's \textcolor{blue}{xxoverxx} 100F''}, but our example \textit{``Save me it’s over 10\textcolor{blue}{2}F''} would pass. 



Many methods have been developed to create text adversarial examples, with the literature predominantly focusing on what we term \newterm{token-modification} attacks. These attacks strategically modify individual document tokens at various granularities (e.g., words, characters) to achieve the desired adversarial effect. (Our above example illustrates a token-modification attack, and Table \ref{table:adversarial_attack} gives more examples.)  In these attacks, the adversary maintains the syntactic structure of the document, unlike, say, a paraphrase or style-transfer model. Token-modification attacks treat the document as an ordered set of tokens and systematically apply \textit{transformations} to individual tokens. These transformations can include operations such as replacing a word with its synonym, deleting a character, or inserting a phrase into the document. The attack process continues iteratively until the adversary achieves its predefined goal or until a specific stopping criterion is met. This results in a controlled and targeted manipulation of the text that can ``fool'' the victim model.


 
Previous surveys, such as those conducted by \citet{WeiEmmaZhangSurvey} and \citet{Alshemali2020}, categorise text attacks based on various criteria: token granularity, adversary goal, model access, targeted NLP task, and so forth. While this categorisation scheme provides a structured overview, it tends to over-emphasise the differences between attacks. In reality, many attack algorithms may only differ in their use of a single component, leading to a situation where algorithms can shift between categories with only trivial modifications. Recognising this limitation, in this survey, we adopt a novel perspective by categorising based on the individual attack components, rather than the attacks themselves. This approach offers a more nuanced understanding of the underlying mechanisms and commonalities across different attacks. To the best of our knowledge, we are the first survey to adopt this component-centric approach.





\input{demonstration_table}

\section{Attack framework}
Recently, \citet{TextAttack} introduced a comprehensive framework that defines four essential components of a textual adversarial attack: (a) a goal function, which determines the success criteria of the attack; (b) a set of transformations, outlining the specific actions that the adversary can undertake; (c) a search method, dictating the sequence in which transformations are applied; and (d) a set of constraints, establishing the boundaries for a valid adversarial example. This formulation casts each text attack as a constrained search problem, defined by its unique combination of these components.

\begin{figure}
\includegraphics[width=\textwidth]{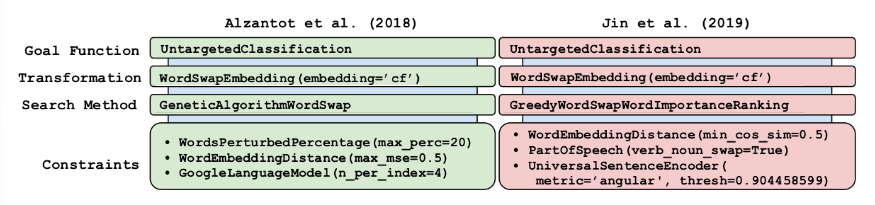}
    \caption{This example, from the TextAttack \citep{TextAttack} documentation, neatly demonstrates how token-modification attacks (\citet{Alzantot2018GA} and \citep{Jin2020TextFooler} here) are made up of a combination of four components. For an easy comparison, the TextAttack paper \citep{TextAttack} also provides a table that summarises the combinations of a number of popular attacks. We cover the individual components in detail in Sections \ref{sec:goal_function}, \ref{sec:transformations}, \ref{sec:search_method}, and \ref{sec:constraints}.}
\label{fig:categorisation}
\end{figure}

Inspired by the work of \citet{TextAttack}, who also used their framework to develop a software library for rapidly implementing text adversarial attacks, we extend this framework and conduct a systematic survey of the various attack components. Our focus is specifically on token-modification attacks, as they represent the predominant attack type in the current landscape. While the framework could theoretically be extended to other attack classes, its application is not straightforward for methods such as those that train a model to generate adversarial examples. Therefore, we restrict our survey to this prevalent and well-defined category of attacks. Our categorisation, along with an exploration of the range and nuances of attack components, is summarised in Figure \ref{fig:categorisation}.






\textbf{Victim model architectures.} The model targeted by an adversarial attack is called the victim model, which in this context are models that perform various NLP tasks. The majority of modern NLP models follow a neural network-based architecture, including convolutional neural networks (CNNs), recurrent neural networks (RNNs), long short-term memory networks (LSTMs), and transformers. This class of models currently achieve the state-of-the-art performance across many NLP tasks and are hence widely used. 

While many adversarial attacks are agnostic to the model's architecture, some techniques assume the victim possesses specific architectural features, such as embedding layers, attention modules, or gradient information. Simpler models, such as a bag-of-words classifier, might not possess these features and the attack will not work. However, it's arguable that these simpler models also inherently possess a multitude of 'adversarial' examples, due to their basic nature and lower performance on legitimate inputs. As such, the assumption is usually made that the victim models targeted are neural network-based.

{\bf Notation.} Let $Ex$ be the original example with label $y_{true}$. All examples are documents, and each document is made up of a set of tokens $\{t_1, \cdots, t_n\}$, for which a token $t_i$ has the numeric representation $\ve_{t_i}$ (e.g. word embeddings). The document $Ex$ is transformed to a numeric representation $\vx$ and then acted on by a model $f$. Likewise, the adversarial example is $Ex'$ (also with label $y_{true}$), its numeric representation is $\vx'$, and we denote its output $f(\vx')$ by $y'$. We use $l$ to represent a generic loss function that acts on a single example.

\section{Adversary goals}
\label{sec:goal_function}

The adversary goal defines what the adversary aims to achieve. Many tasks in NLP, ranging from sentiment analysis and spam detection to more complex tasks like question answering and machine translation, can be broadly classified as having either classification goals or sequence-to-sequence (seq2seq) goals, depending on the task the victim model addresses.

\subsection{Classification}
Attacks on text classifiers, which include tasks such as document categorisation and sentiment analysis, can be broadly considered as either \newterm{targeted} or \newterm{untargeted} attacks. Targeted attacks attempt to induce the model to classify $Ex'$ as a specific class $y_{target}$ different from $y_{true}$ by minimising $l(f(\vx'), y_{target})$. Untargeted attacks only require that $Ex$ is misclassified, and their algorithms typically aim to maximise $l(f(\vx'), y_{true})$. The two are equivalent for binary classification, and for multiclass classification, converting between the two is usually straightforward.\footnote{For example, instead of perturbing $\vx$ towards a \textit{specific} decision boundary, an untargeted attack perturbs $\vx$ towards the \textit{nearest} decision boundary.}

There also exist alternate goal functions. For example, \citet{BadCharacters} propose an \textit{availability attack} goal function that aims to maximise the model processing time.

\subsection{Sequence-to-sequence}
The seq2seq tasks in NLP cover a wide spectrum, including machine translation, text summarisation, paraphrase generation, speech recognition, speech synthesis, and image generation from captions. These tasks often involve complex transformations and mappings between input and output sequences. Unlike classification tasks, where the adversary goals are more standardised, seq2seq tasks do not have a uniform set of adversary goals. Consequently, we adopt a broad classification for seq2seq attacks, categorising them as either untargeted or targeted, to capture the range of possible adversarial objectives within this domain.

\textit{Untargeted attacks:} Untargeted attacks in the seq2seq context aim to degrade the quality of the output sequence as much as possible, as measured by a specific metric. The objective is to disrupt the model's performance without necessarily guiding the output towards a particular target. This form of attack is prevalent in seq2seq tasks and has been employed in various contexts  \citep[e.g.][]{Hsieh2019Robustness,Michel2019Evaluation,Camburu2020MakeUpYourMind,Jia2017Adversarial,Ebrahimi2018CharNMT, PAEG}.

\textit{Targeted attacks:} Targeted attacks, on the other hand, seek to induce specific alterations in the output sequence. The goals can be diverse, ranging from changing a particular output token to any other token \citep{Papernot2016CraftingRNN,Ebrahimi2018CharNMT,Zhao2018GeneratingNatural}, altering every output token from its original value \citep{Seq2Sick}, modifying a set of specific output tokens to another set of specific tokens \citep{Wallace2020Imitation,Ebrahimi2018CharNMT}, to introducing a set of specific output tokens anywhere in the output sequence \citep{Hsieh2019Robustness,Seq2Sick,Zhao2018GeneratingNatural}, or even finding an incorrect input sequence that generates a specific output sequence \citep{Wallace2020Imitation, zhang2021crafting}. These attacks necessitate the careful formulation of an attack-dependent loss function, which is then optimised to either maximise or minimise, depending on the specific adversarial goal.

\section{Transformations}
\label{sec:transformations}
A transformation is a function that alters one ordered set of tokens into another. These transformations can be grouped into several broad categories, referred to here as \textit{transformation types}. Common transformation types include replacing one token with another, inserting a token, deleting a token, changing token ordering, and merging two tokens \citep[e.g.][]{CLARE}. Among these, replacing a token is the most prevalent transformation type, and thus, we give it special emphasis in the following discussion.

When executing a token replacement, where a token $t$ is substituted with a replacement $t_r$, the attack algorithm must select $t_r$ from a \textit{set of replacements}, denoted by $\sS(t)$, with the parameterisation indicating its dependence on $t$. The method for constructing $\sS(t)$ is a core component of many attack algorithms. Typically, the process involves generating an initial set of candidates and then applying constraints to filter the set. For instance, an algorithm might use a dictionary to generate a set of synonyms for $t$ and then filter out candidates that do not share the same part-of-speech (POS) tag with $t$.

Below, we categorise methods that have been employed to generate the initial set of replacement candidates, while deferring the discussion of filtering constraints to Section \ref{sec:constraints}. Attacks can, and do, employ a combination of these methods, and have devised a wide range of nuanced strategies to effectively deceive the victim model.

\input{transforms_table.tex}

\subsection{Any replacement}
The adversary may replace token $t$ with any token of the same granularity. 
While this approach is straightforward and can be highly effective in degrading the victim model's performance, it often results in ungrammatical constructions and loss of semantic meaning. Effectively, these attacks are prioritising minimising the perturbation size without attempting to maintain semantics, and consequently, this method is rarely employed, except in character-level attacks that permit any alphanumeric character (upper or lower case) or punctuation as a valid replacement \citep{Ebrahimi2018HotFlip,Ebrahimi2018CharNMT,Gao2018DeepWordBug, PunctAttack}.



\subsection{Human errors}
The adversary may replace token $t$ with a token that simulates an innocent human-made mistake, such as a typographical error. These transformations leverage the natural imperfections in human writing to craft adversarial examples that blend seamlessly into genuine text, making detection more challenging.

Several transformation types fall under this category:

\begin{itemize}
\item \textit{Typos:} Mimicking keyboard typos by replacing letters with adjacent characters on a QWERTY keyboard \citep{Belinkov2018SyntheticNoise,Textbugger}.
\item \textit{Common misspellings:} Substituting a word with a commonly-misspelled variant, using resources like Wikipedia or corpora tracking document edit history \citep{Belinkov2018SyntheticNoise}.
\item \textit{Number expansion:} Expanding numbers into their word form \citep{BBAEG}.
\item \textit{Leetspeak:} Converting characters to their "leetspeak" equivalents \citep{stylometry2021}.
\item \textit{Other simple transformations:} Examples include removing vowels, shuffling letters, inserting symbols, joining or truncating words, and using phonetic or visually-similar words \citep{zeroe}.
\item \textit{Phonetic similarity:} Choosing replacements that are phonetically similar. This can be done with algorithms like the SOUNDEX algorithm \citep{soundex} or using the Pinyin equivalents for Chinese characters \citep{PerturbationsWild,Liu2023}.
\item \textit{Homograph adjectives:} Homographs are words with the same spelling as another but with a different meaning.  \citet{Emelin2020Detecting} constructed an attack  where they inserted or replaced the adjectives in front of homographs, with the intent to mimic word sense ambiguity errors. 
\end{itemize}

Table \ref{tab:human_errors} provides examples and references for each of these transformation types.

\subsection{Visually similar}
The adversary may replace token $t$ with a visually similar token, intending that a human reader overlooks the discrepancy. These transformations can be applied at the character level, including many-to-one or one-to-many character replacements (e.g., replacing \textit{cl} with \textit{d}). The effectiveness of these attacks may vary depending on the font used and the specific replacements chosen.

Several methods have been proposed to achieve visually similar transformations:

\begin{itemize}
\item \textit{Manual mapping:} Creating a mapping between each token in the vocabulary and a manually chosen replacement \citep{Eger2019VIPER,Textbugger}. For instance, characters may be replaced with versions that include diacritics (e.g., \textit{c} to \textit{č}) \citep{Eger2019VIPER}. Unicode descriptions of characters can also be used to find replacements of the same character and case. In languages like Chinese, characters may be replaced with those sharing partial pictographs \citep{Liu2023}.

\item \textit{Visual embeddings:} Representing each character as a grid of pixels and numerically encoding them, either with raw pixel vectors \citep{Eger2019VIPER} or dense representations from the intermediate layers of a CNN \citep[e.g.][]{Li2020TextShield, UnicodeEvil}. The set of replacements $\sS(t)$ can be constructed by selecting the closest points to the character in the latent space. 

\item \textit{Imperceptible characters:} Designing attacks that are truly imperceptible to the human eye by using invisible characters, homoglyphs, reordering characters, or deletion characters \citep{BadCharacters}. 
\end{itemize}

These visually similar transformations present a unique challenge in adversarial attacks, as they exploit the human tendency to overlook subtle visual discrepancies. The diversity of methods employed reflects the complexity of crafting attacks that are both effective against machine learning models and inconspicuous to human readers.

\subsection{Synonyms}
Adversarial attacks may replace a token $t$ with a synonym that maintains the same meaning. This can be achieved through various methods, including using predefined synonym lists or pre-trained token embedding spaces. Both approaches aim to maintain the semantic content of the original text, while keeping perturbations small and local.

\subsubsection{Predefined list}
The adversary replaces $t$ with a replacement $t_r$ taken from a predefined list of tokens that share the same meaning. Various lists can be used:
\begin{itemize}
\item Thesaurus, lexical database, or knowledge database, such as WordNet \citep{WordNet, Ren2019PWWS}, HowNet \citep{HowNet, Zang2020PSO}, or DrugBank \citep{DrugBank} for biomedical data \citep{BBAEG}.
\item Inflectional perturbations of words with the same base meaning \citep{Tan2020MorphinTime}. 
Inflectional perturbations modify a word's form while retaining its core meaning, such as changing \textit{run} to \textit{running}. 
\item Emoji equivalents for words \citep{TextGuise}.
\end{itemize}
This method often struggles to distinguish the sense in which a word is used, leading to replacements that may read strangely or not fit well in context. Some attacks add constraints to mitigate this issue, but the problem often persists \citep[e.g.][]{Zang2020PSO}.

\subsubsection{Embedding space}
The adversary replaces token $t$ with a token $t_r$ corresponding to a nearby point in a token embedding space. This approach assumes that tokens with similar meanings cluster together in embedding space, forming a natural set of synonyms. 
Different embedding spaces can be used:

\begin{itemize}
\item Using word embeddings like GloVe \citep{GloVe, Textbugger, Xu2020TextTricker}, though this may include related words like antonyms or words with similar concepts but different meanings. Example of antonyms (e.g. \textit{east} and \textit{west}, or \textit{fast} and \textit{slow}) or words with similar concepts but different meanings (e.g. \textit{English}, \textit{American}, and \textit{Australian}).
\item Using \textit{counter-fitted} embeddings \citep{mrksic2016CounterFitting} as a post-processing step for word embeddings. This technique pushes together similar words while also pushing apart related words, resulting in an embedding space where proximity indicates synonymy. It enhances the precision of synonym selection in adversarial attacks \citep{Alzantot2018GA,Jin2020TextFooler,Wang2019IGA,Li2020BertAttack}.
\item Considering contextual embeddings such as BERT \citep{BERT} to provide context-aware token representations. Traditional token embeddings represent a token with a single point, leading to replacements that may not match the token's sense in a specific context. Contextual embeddings like BERT mitigate this issue by considering the surrounding text, enabling more accurate and contextually relevant replacements \citep{Hsieh2019Robustness}.
\end{itemize}

\begin{figure}[h]
    \centering
    \includegraphics[width=0.75\textwidth]{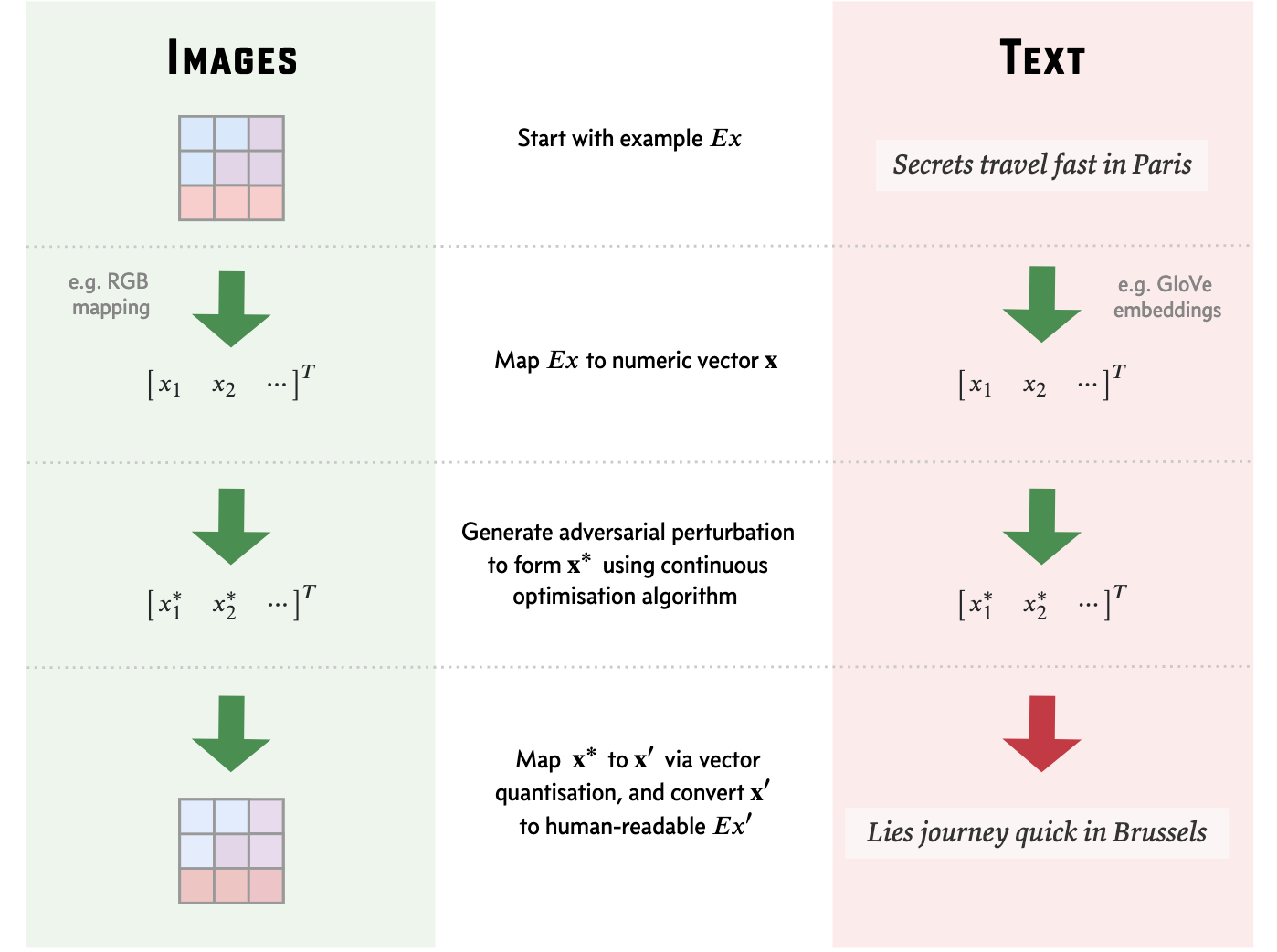}
     \caption[caption]{Continuous perturbations in a numerical space work for image adversarial attacks, where the human eye cannot pick up subtle colour changes, but in the text domain, will be very conspicuous. In other words: while $\vx^*$ fools the model, its text form $Ex'$ may not be semantically valid. The red arrow shows the problematic step.}
    \label{fig:img_vs_text}
\end{figure}

Different methods have been used to define ``closeness'' in embedding space, to identify the set of replacements. Techniques have included:

\begin{itemize}
\item \textit{K-nearest neighbours (KNN):} This approach identifies the $k$ closest tokens to the original in the embedding space, based on distance metrics such as Euclidean distance or cosine similarity \citep[e.g.][]{Textbugger,Xu2020TextTricker}.

\item \textit{Distance thresholding:} Tokens within a specific distance $\delta$ to the original token are considered. The distance can be measured using various metrics, providing flexibility in defining semantic closeness \citep[e.g.][]{Zhang2019generatingfluent}.

\item \textit{Hybrid approach:} Methods can combine the KNN and distance thresholding methods, attempting to ensure that the chosen substitutes are both close and relevant \citep[e.g.][]{Alzantot2018GA,Jin2020TextFooler,Hsieh2019Robustness}.

\item \textit{Synonym clustering:} Creating clusters of synonyms in the embedding space allows for the replacement of a ``central'' word with any word in the cluster \citep{Wang2019IGA}.

\item \textit{Continuous algorithms:} 
In many image-based adversarial attacks, the algorithm iteratively perturbs the input image in a continuous space to cross the decision boundary of the classifier, effectively fooling the model. In the text domain, this concept can be adapted to select the synonyms that move the adversarial example closer to the victim model's decision boundary in the embedding space. For example, \citet{Meng2020geometry} adapted the DeepFool algorithm \citep{DeepFool} for this purpose. Generally, without modification these methods struggle to maintain semantic coherence \citet{Liang2017HotPhrases}, as Figure \ref{fig:img_vs_text} illustrates. \citet{T3} attempt to circumvent this by mapping the text to continuous space and back with tree-based autoencoders and decoders. 

\end{itemize}

\subsection{Language model predictions}
\label{sec:lm}
Language models (LMs) can be used by adversaries to find replacements for tokens, often through \textit{masked language modeling}. The typical process involves replacing the target token with a mask, obtaining predictions from the LM, and using the top predictions as the set of replacements. Variations on this approach are common \citep{Alzantot2018GA,Zhang2019generatingfluent,Li2020BertAttack,Garg2020BAE, CLARE,Lei2022}.
For example, consider the sentence ``\textit{The technology accelerates rapidly}''. The token \textit{accelerates} is masked and then replaced with a top LM prediction, such as \textit{evolves}, resulting in ``\textit{The technology \textcolor{blue}{evolves} rapidly}''. 
The advantages of this approach class include higher-quality replacements, as context is considered, and a tendency to avoid rare tokens, which appear suspicious \citep{Ippolito2020AutomaticDO}. Drawbacks include the need for complex and memory-intensive LMs, slower replacement generation, and the inability to cache synonyms between documents (as they depend on document context). Some approaches mitigate these by combining both LMs and word embeddings \citep{Alzantot2018GA,Zhang2019generatingfluent}. 

Related techniques have included occlusion and language models (OLM) \citep{OLM}; a method from the interpretability literature that identifies relevant words to replace \citep{malik2021advolm}, or Dropout Substitution \citep{DropoutSub}; a technique that applies dropout to internal embeddings of target words to enhance candidate quality \citep{stylometry2021}.

\subsection{Phrase manipulations}
\label{sec:phrase_manipulations}
Attacks have been developed that identify and manipulate whole phrases, rather than tokens. The motivation is to create more linguistically diverse and rich adversarial examples than is possible through a series of individual token-level transforms. We show some examples in Table \ref{tab:phrase_manipulation}. 

To add a phrase into the sentence, a number of approaches can be taken. Another option is to make use of a parse template, which in this context can be thought of as a  predefined format or structure for a sentence. For example, say the adversary wishes to add an adverb phrase to a sentence. The adversary would first identify the examples that it would makes make to add an adverb phrase too, perhaps by comparing the example to the template. Then the adversary would identify an adverb phrase that makes sense in the sentence context, perhaps by using a language model. This is the method used by \citep{AdvExpander}.  

For a controllable phrase replacement process, \citet{Lei2022} propose using two stages to the attack process: identifying phrases in the original sentence suitable to attack, and proposing adversarial replacements that still keep the textual quality. 

To identify the phrase chunks from the original sentence, one approach is to use a syntactic parser, such as the Stanford Parser \citep{StanfordParser}. Parsers like this one identify phrases by first tokenising the text and then tagging each token with its part-of-speech (e.g. noun, verb), dependent on its role in the sentence. It then applies a parsing algorithm to construct a hierarchical syntactic tree that represents the grammatical structure of the sentence. From this tree, phrases can be identified. To identify which phrase to attack, any of the methods in \ref{sec:search_method} can be used. 
 
To find a suitable replacement phrase, a number of methods can be used. A simple approach is to replace individual words with their dictionary definitions, maintaining the same part-of-speech \citep{TextGuise}. While this does add a phrase in, it is easy to introduce a phrase that does not make sense in the overall sentence context. A second approach is to use the mask-infill models described in Section \ref{sec:lm}, where the phrase is masked and then infilled by a language model. The language model chooses the replacement to best suit the context that it is used in the original sentence \citep{Lei2022}. 

\begin{table}[h]
\centering
\begin{tabular}{|p{5.5cm}|p{9cm}|}
\hline
\textbf{Manipulation and Method} & \textbf{Example} \\ \hline
Add a phrase using parse template and language model & Original: ``\textit{Birds fly.}'' \newline Template: ``\textit{Birds fly [ADVP].}'' \newline Adversarial: ``\textit{Birds fly \textcolor{blue}{quickly and quietly}.}'' \\ \hline
Replace a phrase with its dictionary definition & Original: ``\textit{White beaches surrounded by azure seas.} '' \newline Definition: ``azure: bright blue in color like a cloudless sky'' \newline Adversarial: ``\textit{White beaches surrounded by \textcolor{blue}{bright blue in color like a cloudless sky} seas}'' \\ \hline
Replace a phrase using a mask-infill model & Original: ``\textit{The cat sat on the mat.}'' \newline Masked infill: ``\textit{The cat [MASK] the mat.}'' \newline Adversarial: ``\textit{The cat \textcolor{blue}{lay down} on the mat.}'' \\ \hline
\end{tabular}
\caption{Examples of some phrase-level manipulation techniques. }
\label{tab:phrase_manipulation}
\end{table}

\subsection{Set of rules}
One broad approach is to create a set of rules that, when applied, successfully create adversarial examples. These rules are derived from patterns found among successful adversarial examples. \citet{Ribeiro2018} is one example, selecting rules that maximise semantic similarity and have broad applicability. Extending this idea, \citet{yuan2021transferability} crafted rules by identifying adversarial example patterns effective against many different types of models.

\subsection{Homographs}
\citet{Emelin2020Detecting} have devised an attack that makes use of  homographs in the document. Homographs are words that are spelled the same, but have different meanings in different contexts, and also might be pronounced differently. This particular attack targets the words in front of the homograph to subtly change the context of the sentence. For example, the attack might change ``\textit{They were discussing a strike in the meeting.}'' to ``\textit{They were discussing a \textcolor{blue}{lightning} strike in the meeting.}'' Originally, the sentence might suggest a discussion about a labour protest, but the addition of \textit{lightning} changes the context to a weather-related incident. As not every sentence contains targetable homographs in this way, the attack will carefully select examples they can attack, and will leave the rest. 



\section{Search methods}
\label{sec:search_method}
The search method, also referred to as the optimisation method, aims to identify an effective sequence of transformations that satisfies the given constraints of the adversarial problem. The search process terminates either upon achieving the adversarial goal, after a specified number of queries, or when further exploration is restricted by the constraints. 

Search methods typically rank each transformation with a \newterm{scoring function}, also known as an objective or fitness function, that evaluates the quality of the solution in the search space. In the context of adversarial attacks, the scoring function quantifies or estimates the effect of the transformation on the adversarial loss function.

\subsection{Preliminaries}

\subsubsection{Gradient approximations}
The simplest scoring function calculates the increase in loss directly due to the transformation. While exact, this method is computationally expensive because it requires a forward and backward pass through the model for each candidate transformation, which becomes prohibitive for large models.

A widely-used alternative is to use a \textit{gradient-based} scoring function \citep[e.g.][]{Yoo2021Towards, dontsearch}, which employs the gradient of the loss with respect to the input, denoted as $\partial l/\partial \vx$. This gradient serves as a first-order approximation of how the loss $l$ changes when elements of $\vx$ are altered.\footnote{Some attacks \citep[e.g.][]{Papernot2016CraftingRNN} instead use the Jacobian matrix $\partial f(x)/\partial \vx$ and directly change model predictions. }  The transformations are then ranked based on their dot product with the gradient, $\partial l/\partial \vx \cdot (g(\vx) - \vx)$, where $g(\vx) - \vx$ represents the relative change in the input due to the transformation $g$.

This gradient-based approach is faster, requiring only a single backward pass, but it assumes the adversary has access to the model parameters (\textit{white-box} attack). It also rests on two assumptions: that the loss function $l$ is nearly linear around $\vx$, and that other features remain constant when evaluating a single feature's importance \citep{Singla19UnderstandingImpacts}. While these assumptions may not always hold, the method is generally effective in finding successful adversarial examples.

As a compromise, \citet{Wallace2020Imitation} use a hybrid approach which has advantages of both approaches: the speed of the gradient approximation, and the accuracy of direct-checking. They first use the gradient approximation to create a shortlist of the top $k$ transformations across all tokens in $Ex$. They then find the best transformation by directly checking the loss change of each.

\subsubsection{Proxy models} 
In black-box adversarial attacks, where direct access to the target model's architecture or parameters is unavailable, attacks may use proxy models, also known as substitute or replacement models, to aid their attack. The proxy model is intended to mimic the characteristics of the target model, including its gradients, predictions, and confidence values. The attacker usually trains a proxy model on the same dataset as the victim model, or at least, a dataset from a similar domain. The attack can then use the proxy model's gradients or outputs to identify promising replacements for a token, or select where to  attack next. This approach has been used in various works, such as \citep{Papernot2017blackbox, Zhu2022, Yuan2023}.

\subsection{Algorithms}

A variety of optimisation algorithms have been proposed in the literature, which we categorise and present below. 

\subsubsection{Simple heuristics}
The most straightforward is to apply transforms according to a simple heuristic. This is simple to understand but may not find as good a solution as a more advanced algorithm. Some heuristics include

\textit{Batch application:} This method applies a transformation to all eligible tokens in the example $Ex$ at once. Employed by \citet{Tan2020MorphinTime} and \citet{Ebrahimi2018CharNMT}, it replaces each applicable word $w \in Ex$. Variants include multiple runs with each selecting a different transformation \citep{Belinkov2018SyntheticNoise} or replacing each token with a probability $p$ \citep{Eger2019VIPER}.

\textit{Sequential application:} Tokens are transformed in the order they appear in the document. The algorithm iterates through document tokens from start to finish and applying the best transformation for each token until the attack succeeds, cycling if necessary. \citet{Zhang2019generatingfluent} use this method but select transformations based on predefined probabilities, rather than selecting always the best one.

\textit{Random application:} Tokens are randomly selected and transformed. Used as a baseline by \citet{Yoo2020searching} and \citet{Hsieh2019Robustness}, this method replaces tokens either with the best or a random replacement.




\subsubsection{Importance-based methods}
\label{sec:importance}

\begin{figure}
\includegraphics[width=\textwidth]{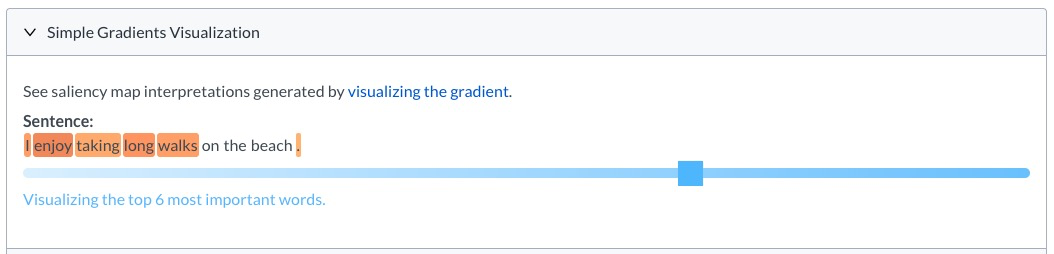}
\caption{A saliency map visualising token importance for a sentiment analysis model. The map ranks the six most impactful words based on their gradient magnitude, shown in shades from dark to light orange. Such visualisations can aid in identifying tokens that, when transformed, are likely to influence model predictions. Obtained from AllenNLP Interpret \citep{AllenNLPInterpret}.}
\label{fig:saliency_map}
\end{figure}

This approach ranks tokens based on their influence on the loss function, transforming the most impactful tokens first to minimise the perturbation size. \citet{Yoo2020searching} term this an \textit{importance ranking}. Saliency maps can be an effective way to visualise these importance rankings, as shown in Figure \ref{fig:saliency_map}.

There are various ways to construct an importance ranking.

\begin{itemize}
\item \textit{Gradient-based ranking:} The gradient magnitude $\lVert \partial l/\partial \ve_{t_i} \rVert_2$ serves as a heuristic for token importance. While this method doesn't capture the direction of the gradient, it remains a widely-used heuristic \citep{Textbugger,Xu2020TextTricker,Samanta2017TowardsCrafting}.
\item \textit{Perturbation-based ranking:} Tokens are systematically perturbed to measure their impact on the loss. Methods include token removal \citep{Jin2020TextFooler,Garg2020BAE,Textbugger} and replacement with neutral tokens like the out-of-vocab, unknown or mask token, or zero vectors \citep{Ren2019PWWS,Ebrahimi2018HotFlip,Li2020BertAttack}. \citet{PARSE} extend this to phrase-level importance.
\item \textit{Temporal metrics:} \citet{Gao2018DeepWordBug} propose metrics based on the document's prefix and suffix relative to each token, termed the \textit{temporal head score} and \textit{temporal tail score}, and that quantify the token's influence on the model's output.
\item \textit{Attention-based ranking:} Attention scores from the victim model's layers are used to rank token importance \citep{Hsieh2019Robustness, maheshwary2021, zeng2021empirical}. This is applicable only to models with attention mechanisms.
\end{itemize}

\citet{Yoo2020searching} found similar effectiveness across these methods,\footnote{They did not test the attention method.}  with gradient-based approaches being more computationally efficient.\footnote{Systematic perturbation methods need many forward passes through the model, but gradient methods only need a single forwards and backward pass.} Subsequent works have proposed additional efficiency improvements, such as using Locality Sensitive Hashing \citep{LSH, maheshwary2021} or predictive networks \citep{guo2022ltg} to reduce the computational burden.

\subsubsection{Beam search and greedy search}
Beam search is a widely-used optimisation technique, often employed in language model decoding, that maintains multiple promising sequences (or ``beams'') simultaneously. The number of sequences is determined by the beam width parameter. A larger beam width generally leads to better solutions but increases computational cost. When the beam width is set to one, the method simplifies to greedy search, which selects the best transformation at each step without considering future implications.

In the context of adversarial attacks, beam search has been employed in works like \citet{Ebrahimi2018HotFlip,Ebrahimi2018CharNMT} and has been reported to achieve high success rates, particularly for shorter documents, while being relatively query-efficient \citep{Yoo2020searching}. Variants of beam search, such as those with added randomness, have also been explored \citep{Zhu2022}. On the other hand, greedy search, used in \citet{Michel2019Evaluation,Wallace2020Imitation}, is computationally less demanding but can get stuck in local minima and may not find the global optimum. \citet{liu2022efficient} introduce a method called local search, which offers a middle ground, and provide theoretical performance bounds for it.



\subsubsection{Population-based algorithms}

Population-based algorithms have been widely used for generating adversarial examples. These algorithms typically evolve a set of candidate solutions over time, guided by a fitness function that evaluates their quality.

\textit{Genetic algorithms:} Among the most prevalent are genetic algorithms, which have been adapted for text-based adversarial attacks by \citet{Alzantot2018GA,Wang2019IGA, hardlabel}. The process begins by generating an initial population of $S$ candidates by applying $S$ different transformations to the original example. In each iteration, new candidates are produced through a series of steps: application of transforms, fitness score calculation based on the loss, parent selection, crossover, and mutation. Different works employ various crossover functions; for instance, \citet{Alzantot2018GA} randomly select words from parent documents, while \citet{Wang2019IGA} splice parent documents at a specific point.


\textit{Differential evolution:} As an alternative to genetic algorithms, differential evolution operates on continuous representations and has been used by \citet{BadCharacters}.


\textit{Particle swarm optimisation (PSO):} PSO has also been applied in this context. The algorithm initialises a swarm of particles, each with a position and velocity. Particles update their positions based on the best-known positions, both locally and globally. They then update their positions based on these best-known positions, with some randomness and a bias towards either the local or global best, depending on the variant.  \citet{Zang2020PSO} used PSO, while \citet{Quantum} explored a variant known as quantum-behaved particle swarm optimisation (QPSO). \citet{Yoo2020searching} noted that PSO is query-inefficient despite its effectiveness in finding good solutions.



\subsubsection{Other}
Various other search methods have been used. These have included 

\begin{itemize}
\item \textit{Learning-based approaches:} These methods integrate machine learning algorithms directly into the optimisation process. For instance, \citet{stylometry2021} use neural networks to identify which tokens in the text are most susceptible to perturbation. Similarly, \citet{uap} use a reinforcement learning algorithm to choose synonym swaps. 

\item \textit{Minimising perturbation size:} Some approaches focus on generating adversarial examples with small perturbations to the original text. \citet{rr} introduce a rollback operation that iteratively removes unnecessary edits, thereby reducing the perturbation size.  In a similar vein, \citet{Lee2022Query} use a Bayesian optimisation technique that is further refined using importance scores and a post-processing step that minimises the perturbation size.
\end{itemize}






\section{Constraints}
\label{sec:constraints}

As mentioned above, adversarial attacks all operate under a set of constraints. A constraint in this context refers to a rule or set of rules that determine what is an acceptable adversarial example. These constraints generally ensure that the perturbed input, while still being effective in misleading the classifier, remains as close as possible to the original input in terms of semantics, structure, and readability. For instance, constraints may dictate that the modified text should not deviate significantly in meaning from the original example (semantic drift), must retain readability for human observers, should adhere to grammatical norms, and maintain a certain level of similarity between the original and adversarial examples. These constraints preserve the linguistic integrity of the adversarial samples and ensure that the attacks are realistic and plausible within the bounds of the assumed threat model. 

As constraints will control the quality and characteristics of the generated adversarial examples, setting weaker linguistic constraints will increase the success rate of attacks, as the adversary has more freedom to alter the input text.
However, the resultant adversarial examples will have lower text quality, which in this context, means they may unnatural or easily identifiable as adversarial interventions.  Conversely, stronger constraints will produce higher quality adversarial examples that more closely resemble legitimate inputs. Yet, these constraints can also make it more challenging to find successful perturbations, and will lower the attack's success rate. This trade-off is why attack success rate cannot be used to compare attack algorithms with different constraints, as the attack with weaker constraints will perform better \citep{Yoo2020searching}. 

In the following subsections, we categorise and present the different constraints that have been used in the literature. Table \ref{tab:constraints} provides a top-level overview.

\input{constraint_table}

\subsection{Readability}
These constraints are intended to enforce that the adversarial example be as close to natural text as possible. 

\paragraph{Grammar}
A simple constraint in this area is that the adversarial example be grammatical. The intention is that adversarial examples that are more grammatical will naturally be of higher quality. This constraint can be checked using grammar-checking software, such as LanguageTool \citep{LanguageTool}, as done by \citet{Lei2022}). However, this constraint is not applicable to all attacks or threat models. It wouldn't make sense when creating adversarial examples on data from domains which are rarely grammatical, such as tweets or forum posts. For example, the attack by \citet{Tan2020MorphinTime} is aimed to deliberately mimic non-native English speakers, which makes grammar-checking less relevant in their context. Here, a grammatically correct sentence, ``\textit{The cat sat on the mat}'' could be transformed to ``\textit{Cat sat on mat}'', which in this context is a legitimate adversarial example.

\paragraph{Fluency}
Attacks which value fluency wish to produce smooth, naturally flowing text that is easily comprehensible and linguistically sound. These attacks preference fluent sentences, like ``\textit{The quick brown fox jumps over the lazy dog}'', over non-fluent ones, such as ``\textit{Quick brown fox jump over lazy dog}''. 

The most common approach to measure fluency is to score the text by computing its likelihood according to a pre-trained language model, with the intention that more fluent text will score higher than less fluent text. Scoring methods range from the simple measures of perplexity and entropy, to developed metrics like BARTScore \citep{BARTScore} or BERTScore \citep{BERTScore}. These methods, akin to using one generative model to evaluate another \citep{deutsch-etal-2022-limitations} are widely used, often as a heuristic to select token replacements \citep{Alzantot2018GA,han2020structured,Quantization}. 

An alternate approach is to assess the `linguistic acceptability' of the text \citep{TextAttack}. Linguistic acceptability is the task of determining if a sentence is acceptable or not, as judged by native speakers \citep{CoLA}. This is evaluated using a model that has been trained on a linguistic acceptability dataset, such as the Corpus of Linguistic Acceptability \citep{CoLA}, and that then assesses text as either acceptable or unacceptable. Fluent text is more likely to be marked as acceptable because of its preservation of natural grammatical structures. As an example from this dataset, ``\textit{The dog barked its way out of the room.}'' would be acceptable, but ``\textit{The dog barked out of the room.}'' would be unacceptable.

\paragraph{Character replacement constraints}
Certain constraints are specifically tailored for character-level attacks --- the attacks that add, remove or replace individual characters to achieve success. These attacks often introduce spelling mistakes and typos. Hence, a constraint is to require that the adversarial example contain no spelling errors, according to a spellchecker. To improve readability, \citet{Belinkov2018SyntheticNoise} have enforced additional constraints on what character transformations are acceptable. They disallow character replacements for words less than a certain length, and require that the first and last letters of each word are preserved. Examples of successful modifications under all these criteria might be changing \textit{bead} to \textit{bed}, or \textit{house} to \textit{horse}.

\subsection{Semantic}
\label{sec:semantic}
Semantic constraints aim to ensure that the adversarial example preserves the original document's intent and meaning. These constraints are essential because in text, a single token change can significantly change the document's meaning. Consider the sentence: "\textit{The doctor recommended regular exercise to improve health.}"  Here, the word ``\textit{exercise}'' is a key component of the sentence. If this word is changed to ``\textit{medication}'' the sentence becomes  ``\textit{The doctor recommended regular \textcolor{blue}{medication} to improve health}'', and the entire meaning of the sentence has shifted dramatically.

We can divide semantic constraints into a number of categories:

\paragraph{Token selection rules}
Constraints in this category govern how tokens in the original document can be transformed.

One type of constraint specifies which tokens can be acted upon, aiming to avoid changing tokens that are vital for document structure. For instance, some methods forbid the replacement of stop words \citep{Jin2020TextFooler,Xu2020TextTricker}, or using stop words as a replacement \citet{Li2020BertAttack}. Stop words are commonly used words in a language that contribute to the grammatical structure but carry little semantic weight, such as ``\textit{the}'', ``\textit{is}'', ``at'', and so on. For example, modifying stop words in a sentence like ``\textit{She is reading a book}'' to ``\textit{She \textcolor{blue}{of} reading a book}'' significantly degrades the sentence's quality and changes its interpretation. A version of this constraint only specifically allows the replacement of specific word types, such as nouns, verbs, adverbs, or adjectives\citep{Tan2020MorphinTime,Zang2020PSO}.

Another common constraint is to allow only one transformation per token \citep{Seq2Sick,Ebrahimi2018HotFlip}, or more generally, up to \(k\) transformations per token \citep{Wang2019IGA}. This helps to minimise potential semantic drift caused by multiple token changes. For example, in the sentence ``\textit{The quick brown fox jumps over the lazy dog}'', say the attacker replaces the second token, \textit{quick}, with a synonym four times. The token changes from \textit{quick} to \textit{swift} to \textit{immediate} to \textit{instantaneous}, and the adversarial example ``\textit{The \textcolor{blue}{instantaneous} brown fox jumps over the lazy dog}'' has lost the meaning of the original. 

Some token selection rules are specific to a particular NLP task. One example is for sentiment analysis attacks, where constraints may forbid the replacement of sentiment-expressing to avoid changing the ground-truth label of the sentence.  For example, if the original sentence is `\textit{`I am extremely happy with the service}'', the constraints would prevent changing the sentiment-expressing word ``happy''. Modifying negation words like ``not'' could also be forbidden. These constraints are seen in \citet{Zhang2019generatingfluent}.

\paragraph{Token replacement rules}
Constraints in this category aim to ensure that, when replacing one token with another, the replacement is similar to the original in the context of the document.

One approach requires that replacements share the same part-of-speech tag as the original token, as seen in approaches by \citet{Jin2020TextFooler,Zang2020PSO,Xu2020TextTricker}. This helps to eliminate potential synonyms that use an incorrect word sense. Consider the sentence ``\textit{She was absolutely ecstatic about her results.}'' The constraint would allow the adversary to change \textit{ecstatic} (adjective) to \textit{elated} (adjective), forming ``\textit{She was absolutely \textcolor{blue}{elated} about her results.}'', but would prohibit changing \textit{ecstatic} (adjective) to \textit{ecstasy} (noun), and then forming ``\textit{She was absolutely \textcolor{blue}{ecstasy} about her results.}'' 

Another approach by \citet{Lei2022} attempts to maintain label preservation of the original class by evaluating replacements using a class-conditioned language model \citep{CCLM}. This is a pre-trained language model that is then fine-tuned on only examples of a particular class of the dataset, such as positive examples for a sentiment analysis dataset. If, after a token replacement, a text has a high likelihood according to one of these models trained on the class of the original example, the reasoning is that it must belong to a similar distribution as the training data --- and hence, this text should be truly labelled as of original class. 

\paragraph{Document embeddings}
Document embeddings represent entire documents as fixed-size, dense numerical vectors. They allow the rich and complex information contained in a document into a format that is easily interpretable by machine learning models. An earlier way to create them was to aggregating individual word embeddings, such as those from GloVe \citep{GloVe}. A more modern method would be to use a transformer, (e.g. T5 \citep{T5}) to process the document and then pool the output embeddings. Adversarial attacks have tended to use the Universal Sentence Encoder \citep{UniversalSentenceEncoder} (USE), a model designed to produce sentence embeddings; examples include \citet{Liu2023} or \citet{Jin2020TextFooler}. A similar model used for other NLP tasks is the Sentence-BERT model \citep{reimers-gurevych-2019-sentence}, which modifies the BERT architecture to produce embeddings more suitable to semantic similarity tasks. 

Document embeddings operate on the principle that documents with similar meanings should have closely related embeddings. To evaluate the similarity between the original document and its adversarial counterpart, their embeddings can be compared. This comparison is often done using cosine similarity. If the similarity score is too low (below a certain threshold), it indicates a significant deviation in meaning, and as a result, the proposed transformation of the text is either avoided \citep{Jin2020TextFooler} or given a lower priority \citep{Garg2020BAE}.




\paragraph{Distance limits} 
In many adversarial attacks on images (e.g. \citep{FGSM}) it is common to put a constraint on the maximum distance between the original and and adversarial example, typically measured with a \(L_p\) norm.  This constraint is intended to ensure that the adversarial changes are imperceptible to the human eye, and the image appears unaltered. This concept of limiting the adversarial distance has been similarly applied in text-based adversarial attacks. As imperceptible changes are not possible, perhaps the equivalent goal here is to keep the adversarial text semantically and structurally close to the original, so that the changes are not so obvious to humans.

A common distance constraint is a maximum edit distance limit. Edit distance measures the number of transformations needed to change one text to another. One variant is the Hamming distance, which measures the minimum number of token replacements needed to convert between the original and adversarial example. Another is the Levenshtein distance, which also allows inserting and deleting tokens, on top of replacement. For instance, consider again the sentence ``\textit{The quick brown fox jumps over the lazy dog.}'' If we change \textit{quick} to \textit{swift} and \textit{lazy} to \textit{slow} the Hamming distance would be two, for the two token replacements. On the other hand, if we change \textit{quick} to \textit{swift}, remove \textit{lazy} and insert \textit{sleepy} before \textit{dog}, the Levenshtein distance would be three, for one replacement, one deletion, and one insertion. Edit distance is commonly used \citep[e.g.][]{Ebrahimi2018CharNMT,Gao2018DeepWordBug,Textbugger}

Other distance metrics could include n-gram based metrics: those that track the number of matching n-grams between the original and adversarial examples. One example of this type of metric is BLEU \citep{BLEU}, commonly used in machine translation. Another proposed option is  Word Mover Distance \citet{WeiEmmaZhangSurvey}, which broadly speaking, tracks the minimum distance that words from one text need to `travel' in embedding space to match exactly with the words in another document. However, these metrics have not been widely adopted.

\subsection{Performance}
Some constraints are intended to improve the attack's  success rate, rather than to necessarily increase the text quality. Below are some examples of these.  

Some approaches attempt to exploit the special tokens of the tokeniser to fool the victim model. For example, \citet{Xu2020TextTricker} include padding and unknown tokens in the synonym set of words to increase the variety of potential replacements. This allows adversarial examples like ``The \textcolor{blue}{[PAD]} brown fox jumps over the lazy dog''.  Another approach along these lines is to allow character-level transformations only if the resulting word is not in the model's vocabulary, as done by \citet{Ebrahimi2018HotFlip}. This is done so that victim models that tokenise on the word level will then represent the changed word with an out-of-vocabulary token, rather than with a different, valid word token.  This can lead to more effective adversarial examples, as the model cannot leverage its understanding of the altered word. For example, if the sentence was ``\textit{I cannot sit down on this chair.}'', the adversary could change \textit{chair} to \textit{chbir}, but not to \textit{chain}. Then, \textit{chbir} would be replaced at tokenisation with the out-of-vocabulary token. 

To improve the search space of the attack, \citet{Zang2020PSO} uses lemmatisation when searching for synonyms, and delemmatisation afterwards to prevent grammatical mistakes. Lemmatisation reduces words to their base or root form (lemma), while delemmatisation reverts lemmas back to their original inflected forms. For example, consider the sentence ``\textit{The cats are sleeping on the mat.}'' With lemmatisation, \textit{cats} would be reduced to its base form \textit{cat} and similarly, \textit{sleeping} to \textit{sleep}. Then synonyms are searched for at the lemma level, which allows for a broader set of synonyms.  After synonyms are found, like \textit{feline} for \textit{cat} and \textit{rest} for \textit{sleep}, delemmatisation is applied to fit these synonyms back into the original sentence context. The final sentence would then be ``\textit{The \textcolor{blue}{felines} are \textcolor{blue}{resting} on the mat.}''

To increase attack strength, some attacks constrain replacements to those commonly seen in examples of other classes.  For example, \citet{Ren2019PWWS} replace named entities with the most frequent same-type named entities from a different class. Named entities are specific, identifiable elements in text, such as names of people or places, that are categorised as distinct types through the task of named entity recognition. In this approach, named entities in a text are replaced with other named entities of the same type but often seen with examples of a different class to the original in the training dataset. For example, consider the sentence, ``\textit{I bought a smartphone from Apple.}'' Here, \textit{Apple} is a named entity, categorised as an organisation. With this strategy, \textit{Apple} might be replaced with another organisation, such as \textit{Samsung}, if \textit{Samsung} is identified as a frequent entity for a different class.  Similarly, \citet{Samanta2017TowardsCrafting} categorise each example (e.g., by movie genre for a dataset of movie reviews) and then prioritise word replacements that are frequently used both within that category and in examples from different classes.



\section{Conclusion}
\label{sec:future_directions}
In this survey, we have provided an overview of the key components that make up token-modification adversarial attacks: goal functions, transformations, search methods, and constraints. 

We have first covered \textit{goal functions}, which define what the adversary would like to achieve, and allow for a quantifiable measure of progress towards that goal. We have classified these as either classification or seq2seq goal functions, and again as either targeted or untargeted attacks, depending on the NLP task and the specifics of the attack goal. The majority of adversarial attacks have been for text classification, and it would be interesting to see more attacks targeting the wide range of seq2seq tasks, such as dialogue systems, paraphrasing, text summarisation and code generation. 


We have then covered the wide range of proposed \textit{transformations}: functions that alter one set of ordered tokens into another, where the token can be at a range of different granularities. Typical transformations insert, replace or delete tokens, with most attacks replacing and fewer inserting or deleting them. The type of transformation depends on the intent of the adversary. If they simply wish to induce a label flip for a classifier, they can change tokens as they like. If they wish to mimic human errors, they might introduce typos or phonetically similar words to the text. If they wish to create close-to-imperceptible adversarial examples, they may replace with visually similar tokens or use invisible characters. If they wish to preserve the fluency of the text, they might use a language model to mask a word, and then replace it --- and so on. 

After, we covered \textit{search methods}. Starting from an original example, the search method repeatedly applies transformations until the adversarial example is found, or the search fails. They judge and choose transformations based on scoring functions, which numerically rank the available options. Search methods may be allow access to the gradients of the victim model (white-box) or they may forbid access (black-box). Should the victim model not be available, search methods may use a substitute model instead. A variety of search methods have been proposed. Simple methods and heuristics, such as greedy search, run quickly but give worse results. More complex methods, such as particle swarm optimisation, give better results but are slower to run.  Generally, there is a correlation between the number of times the victim model is queried, and the success of the search method; \citet{Yoo2020searching} has provided an initial benchmark of the performance and query effectiveness of search methods. More efficient search methods are continuing to be identified \citep[e.g.][]{Lee2022Query,liu2022efficient, guo2022ltg}.

We have also surveyed the \textit{constraints} of the field. Constraints determine what makes successful or unsuccessful adversarial examples. They are used to enforce minimum standards on readability, fluency, grammatically and semantic similarity, or also to set a maximum allowed perturbation distance. Constraints can also increase the attack strength, such as to exclude transformations less likely to degrade the performance of the victim model. Constraints are often quantified using automated metrics, like the count of grammatical errors or language model entropy for assessing fluency. These constraints are generally employed at each stage of the optimisation algorithm to prevent the search process from deviating.



Finally, we will now propose some future directions that are motivated by the limitations of the field.

\paragraph{Standardise attack scenarios} Currently the field is limited by a lack of standardised attack scenarios. The first problem is that each study establishes its own threat model, transformations, and constraints, which does not easily allow for cross-study comparisons. The second problem is that most papers' threat models do not align with real-world attack scenarios, which limits their practical relevance \citep{Gilmer2018rules}.  Real-world scenarios might include, among others: circumventing email spam filters; manipulating user reviews to deceive sentiment analysis models; causing errors in translation models; altering news articles to evade fact-checking models; challenging hate-speech, offensive-language, or other automated content moderation systems; altering documents to evade piracy detection; or prompting chatbots to provide incorrect advice.

\paragraph{Extend to multimodal scenarios} This has been a survey of text attacks, but it is worth remembering that many real-world systems are multimodal,  integrating text, images, videos, audio, graphs, spatial data, structured data, time-series data, and more. For instance, webpages and emails often contain a mix of text, hyperlinks, images, video, and audio. Navigation systems meld text with images and geospatial data. A Facebook account may include a profile bio, a feed of text, video, and image posts, a network of friends, and geographical details (e.g., country of origin). Adapting text adversarial attack techniques to multimodal systems is a promising direction for future research. 

\paragraph{More multiple-token attacks} Most attacks insert, delete or replace single tokens at each step, which significantly limits the range of possible adversarial examples. By contrast, in Section \ref{sec:phrase_manipulations} we discussed phrase-level attacks, such as the attacks by \citet{rr} and \citet{Lei2022} that alter multiple tokens per optimisation step. These attacks are capable of a much larger range of adversarial examples, which is appealing, but to date they are much less explored. A promising option could be to incorporate generative seq2seq models, such as paraphrase models, to identify high-quality phrase and sentence-level transforms. Such an attack could modify phrases or idioms, which single-token attacks could not do. Consider the idiom ``\textit{Bite the bullet}'', which refers to enduring a painful or unpleasant situation.  Such an attack might replace it with a similar idiom, such as ``\textit{Grin and bear it}'' or ``\textit{Tough it out}'', which is not easily achievable with single-token modification algorithms. 


%

\paragraph{Incorporate human judgements}
Currently attacks are often judged and ranked by automated metrics, such as fluency or semantic similarity scores.  These metrics are flawed, however, and often correlate poorly with human judgement \citep{deutsch-etal-2022-limitations, Morris2020reeval}, leading to scenarios where many so-called `adversarial' examples are non-fluent, ungrammatical, do not preserve semantics, change the ground-truth label, etc. \citep{herel2022preserving}. The core problem is that it is very hard to capture all the required `adversarial' properties with a collection of metrics, as the ultimate judge of adversarial attacks is a human, and the metrics deviate from human judgement. Directly incorporating human feedback would improve the quality of adversarial examples, but human judgements are typically difficult and expensive to obtain.  

Two alternatives come to mind. The first is to allow the end user to construct the adversarial example, perhaps with guidance from a user-friendly interface and assistance from a model, as demonstrated in \citep{Wallace2019Quizbowl}.  The second is to use techniques like Reinforcement Learning from Human Feedback (RLHF).  Here, an adversary would develop a 'reward' model that assigns scalar rewards to assess the quality of adversarial examples. The reward model would be trained using human preferences, and would be applied as a constraint in the optimisation process.  The use of RLHF was key to the success of ChatGPT and other similar systems. It is particularly useful for scenarios where defining an explicit reward function is difficult, such as for this one.

\backmatter





\section*{Statements and declarations}

\textbf{Authorship statement.} All authors contributed to the conception and design of the paper. The first author, Tom Roth, performed the literature search and wrote the manuscript. All authors then critically revised the work, commented on previous versions, and read and approved the final manuscript.

\noindent\textbf{Declarations.} This research was supported by scholarships from Data61 and from the Australian Government Research Training Program (RTP). The authors declare that they have no conflict of interest and that they have no other sources of funding.  

\noindent\textbf{Data availability statement.} Data sharing is not applicable to this article as no datasets were generated or analysed during the current study.



\bibliographystyle{ios1}           
\bibliography{bibliography}        

%

\end{document}

%% file: demonstration_table.tex
\begin{table}[h]
\centering
\begin{tabular}{|c|l|}
\hline
\textbf{Granularity} & \textbf{Example} \\
\hline
Original & \textit{The movie was absolutely fantastic and I loved it!} \\
\hline
Character-level & \textit{The movie was absolutely f\textcolor{blue}{4}ntastic and I loved it!} \\
\hline
Subword-level & \textit{The movie was absolutely fantast\textcolor{blue}{ical} and I loved it!} \\
\hline
Word-level & \textit{The \textcolor{blue}{film} was absolutely fantastic and I loved it!} \\
\hline
Phrase-level & \textit{The \textcolor{blue}{cinematic experience} was absolutely fantastic and I loved it!} \\
\hline
\end{tabular}
\caption{Another example of a token-modification adversarial attack on a sentiment classifier. The original example is categorised as positive, and the adversarial attack attempts to change the predicted label to negative through changing its tokens, which can be at a range of granularities. }

\label{table:adversarial_attack}
\end{table}

%% file: transforms_table.tex
\begin{table}[h]
\small
\centering
\begin{tabular}{|l|l|l|}
\hline
\textbf{Transform type} & \textbf{Example} & \textbf{Sample references} \\ \hline
Any token replacement & \textit{dog} $\rightarrow$ \textit{!og} & \citep{Ebrahimi2018HotFlip,Ebrahimi2018CharNMT,Gao2018DeepWordBug,PunctAttack} \\ \hline
Adjacent keyboard typos & \textit{achieve} $\rightarrow$ \textit{axhieve} & \citep{Belinkov2018SyntheticNoise,Textbugger} \\ \hline
Common misspellings & \textit{achieve} $\rightarrow$ \textit{acheive} & \citep{Belinkov2018SyntheticNoise} \\ \hline
Number to word form & \textit{12} $\rightarrow$ \textit{twelve} & \citep{BBAEG} \\ \hline
Leetspeak conversions & \textit{leet} $\rightarrow$ \textit{l33t} & \citep{stylometry2021} \\ \hline
Vowel removal, letter shuffling, etc  & \textit{letters} $\rightarrow$ \textit{ltetrs} & \citep{zeroe} \\ \hline
Phonetic replacements & \textit{fair} $\rightarrow$ \textit{fare} & \citep{PerturbationsWild,Liu2023} \\ \hline
Homograph adjective manipulation & \textit{short lead} $\rightarrow$ \textit{hot lead} & \citep{Emelin2020Detecting} \\ \hline
Unicode-based replacements & \textit{a} $\rightarrow$ \textit{à} & \citep{Eger2019VIPER} \\ \hline
Visual character mapping & \textit{cl} $\rightarrow$ \textit{d} & \citep{Li2020TextShield, UnicodeEvil} \\ \hline
Invisible characters & \textit{abc} $\rightarrow$ \textit{abc<U+200B>} & \citep{BadCharacters} \\ \hline
Inflectional perturbations & \textit{run} $\rightarrow$ \textit{running} & \citep{Tan2020MorphinTime} \\ \hline
Thesaurus-based synonyms & \textit{happy} $\rightarrow$ \textit{joyful} & \citep{Ren2019PWWS,WordNet,Zang2020PSO} \\ \hline
Word to emoji & \textit{love} $\rightarrow$ \heart & \citep{TextGuise} \\ \hline
Word embedding-based & \textit{fast} $\rightarrow$ \textit{swift} & \citep{GloVe,Textbugger,Xu2020TextTricker} \\ \hline
Counter-fitted synonyms & \textit{happy} $\rightarrow$ \textit{content} & \citep{mrksic2016CounterFitting,Alzantot2018GA} \\ \hline
Context-aware embeddings & \textit{bank} $\rightarrow$ \textit{credit union} & \citep{Li2020BertAttack} \\ \hline
Masked language model infill & \textit{go [MASK](here) fast} $\rightarrow$ \textit{go very fast} & \citep{Alzantot2018GA,Zhang2019generatingfluent,Li2020BertAttack,Garg2020BAE, CLARE,Lei2022} \\ \hline
Phrase replacements & \textit{quickly run} $\rightarrow$ \textit{swiftly sprint} & \citep{Lei2022,AdvExpander,TextGuise} \\ \hline
Predefined parse template filling & \textit{[Noun] runs [Adverb]} $\rightarrow$ \textit{The dog runs quickly} & \citep{AdvExpander} \\ \hline
Derived rule-based & \textit{What NOUN} $\rightarrow$ \textit{Which NOUN} & \citep{yuan2021transferability} \\ \hline
\end{tabular}
\caption{Examples of different transformation types, and some examples of methods where they were used. }
\label{tab:human_errors}
\end{table}

%% file: constraint_table.tex
\begin{table}[h]
\centering
\begin{tabular}{|l|l|}
\hline
\textbf{Description} & \textbf{Sample references} \\
\hline
Apply spellchecker& \citet{Belinkov2018SyntheticNoise} \\
\hline
Limit character transformations to words of specific length & \citet{Belinkov2018SyntheticNoise} \\
\hline
Keep first and last letters of each word & \citet{Belinkov2018SyntheticNoise} \\
\hline
Require pass through grammar checker  & \citet{Lei2022} \\
\hline
Check with linguistic acceptability model & \citet{TextAttack} \\
\hline
Fluency via language model perplexity & \citet{wang2021closer} \\
\hline
Max of one transformation per token & \citet{Seq2Sick,Ebrahimi2018HotFlip} \\
\hline
Max of \(k\) transformations per token & \citet{Wang2019IGA} \\
\hline
No replacing stop words & \citet{Jin2020TextFooler,Xu2020TextTricker} \\
\hline
Only replace specific word types (e.g. noun) & \citet{Tan2020MorphinTime,Zang2020PSO} \\
\hline
Disallow stop words as replacements & \citet{Li2020BertAttack} \\
\hline
Same part-of-speech tag for replacements & \citet{Jin2020TextFooler,Zang2020PSO,Xu2020TextTricker} \\
\hline
Filter replacements by fluency constraint & \citet{Alzantot2018GA,han2020structured,Quantization, wang2021closer} \\
\hline
Attack only sentences with homographs & \citet{Emelin2020Detecting} \\
\hline
Score replacements with class-conditioned language model & \citet{Lei2022} \\
\hline
Min cosine similarity between document encodings & \citet{Jin2020TextFooler,Garg2020BAE,UniversalSentenceEncoder,Liu2023} \\
\hline
Forbid replacement of sentiment or negation words & \citet{Zhang2019generatingfluent} \\
\hline
Min similarity on round-trip translation & \citet{zhang2021crafting} \\
\hline
Max edit distance & \citet{Ebrahimi2018CharNMT,Gao2018DeepWordBug,Textbugger} \\
\hline
Include padding and unknown tokens in synonym set & \citet{Xu2020TextTricker} \\
\hline
Use lemmatisation and delemmatisation for candidate synonyms & \citet{Zang2020PSO} \\
\hline
Allow character transformations only for out-of-vocab words & \citet{Ebrahimi2018HotFlip} \\
\hline
Prioritize word replacements based on class and frequency & \citet{Samanta2017TowardsCrafting, Ren2019PWWS} \\
\hline
\end{tabular}
\caption{Summary of the different constraints used in adversarial attacks.}
\label{tab:constraints}
\end{table}